\pdfoutput=1

\documentclass[11pt]{article}

\usepackage[]{acl}

\usepackage{times}
\usepackage{latexsym}
\usepackage{graphicx}

\usepackage[T1]{fontenc}

\usepackage[utf8]{inputenc}

\usepackage{microtype}

\usepackage{inconsolata}
\usepackage{booktabs}
\usepackage{enumitem}
\usepackage{bbm}
\usepackage{multirow}
\usepackage{xspace}
\usepackage{xcolor}
\usepackage{colortbl}
\usepackage{float}
\usepackage{tcolorbox}

\DeclareUnicodeCharacter{0307}{\.}
\newcommand{\dataset}{\textsc{InstruSum}\xspace}

\newtcolorbox{promptbox}[2][Prompt]{
colback=black!4!white,
arc=5pt,
boxrule=1.1pt,
fonttitle=\bfseries,
title=#1,
before upper={\small},
fontupper=\selectfont\footnotesize,
colframe=#2,
}

\title{Benchmarking Generation and Evaluation Capabilities of Large Language Models for Instruction Controllable Summarization}

\author{
 Yixin Liu\Thanks{~~Equal contribution} $^{1}$ 
 \quad \textbf{Alexander R. Fabbri}\textsuperscript{$*$}$^{2}$
 \quad \textbf{Jiawen Chen}$^{1}$
 \quad \textbf{Yilun Zhao}$^{1}$
 \quad \textbf{Simeng Han}$^{1}$ \\
 \quad \textbf{Shafiq Joty}$^{2}$ 
  \quad \textbf{Pengfei Liu}$^{3}$ 
 \quad \textbf{Dragomir Radev}$^{1}$
 \quad \textbf{Chien-Sheng Wu}$^{2}$ 
 \quad \textbf{Arman Cohan}$^{1,4}$ \vspace{6pt}\\
  $^1$Yale University\quad 
  $^2$Salesforce AI
  \quad
  $^3$Shanghai Jiao Tong University 
  \quad
  $^4$Allen Institute for AI
  \vspace{6pt}\\
  \texttt{yixin.liu@yale.edu, afabbri@salesforce.com, arman.cohan@yale.edu}
 }

\begin{document}
\maketitle

\begin{abstract}
While large language models (LLMs) can already achieve strong performance on standard generic summarization benchmarks, their performance on more complex summarization task settings is less studied.
Therefore, we benchmark LLMs on \textbf{\textit{instruction controllable}} text summarization, where the model input consists of both a source article and a natural language requirement for desired summary characteristics. 
To this end, we curate an evaluation-only dataset for this task setting and conduct human evaluations of five LLM-based systems to assess their instruction-following capabilities in controllable summarization.
We then benchmark LLM-based automatic evaluation for this task with 4 different evaluation protocols and 11 LLMs, resulting in 40 evaluation methods.
Our study reveals that instruction controllable text summarization remains a challenging task for LLMs, since 
(1) all LLMs evaluated still make factual and other types of errors in their summaries;
(2) no LLM-based evaluation methods can achieve a strong alignment with human annotators when judging the quality of candidate summaries;
(3) different LLMs show large performance gaps in summary generation and evaluation capabilities.
We make our collected benchmark \dataset publicly available to facilitate future research in this direction.

\end{abstract}

\section{Introduction}

\begin{figure}[t!]
    \centering
    \includegraphics[width=0.85\linewidth]{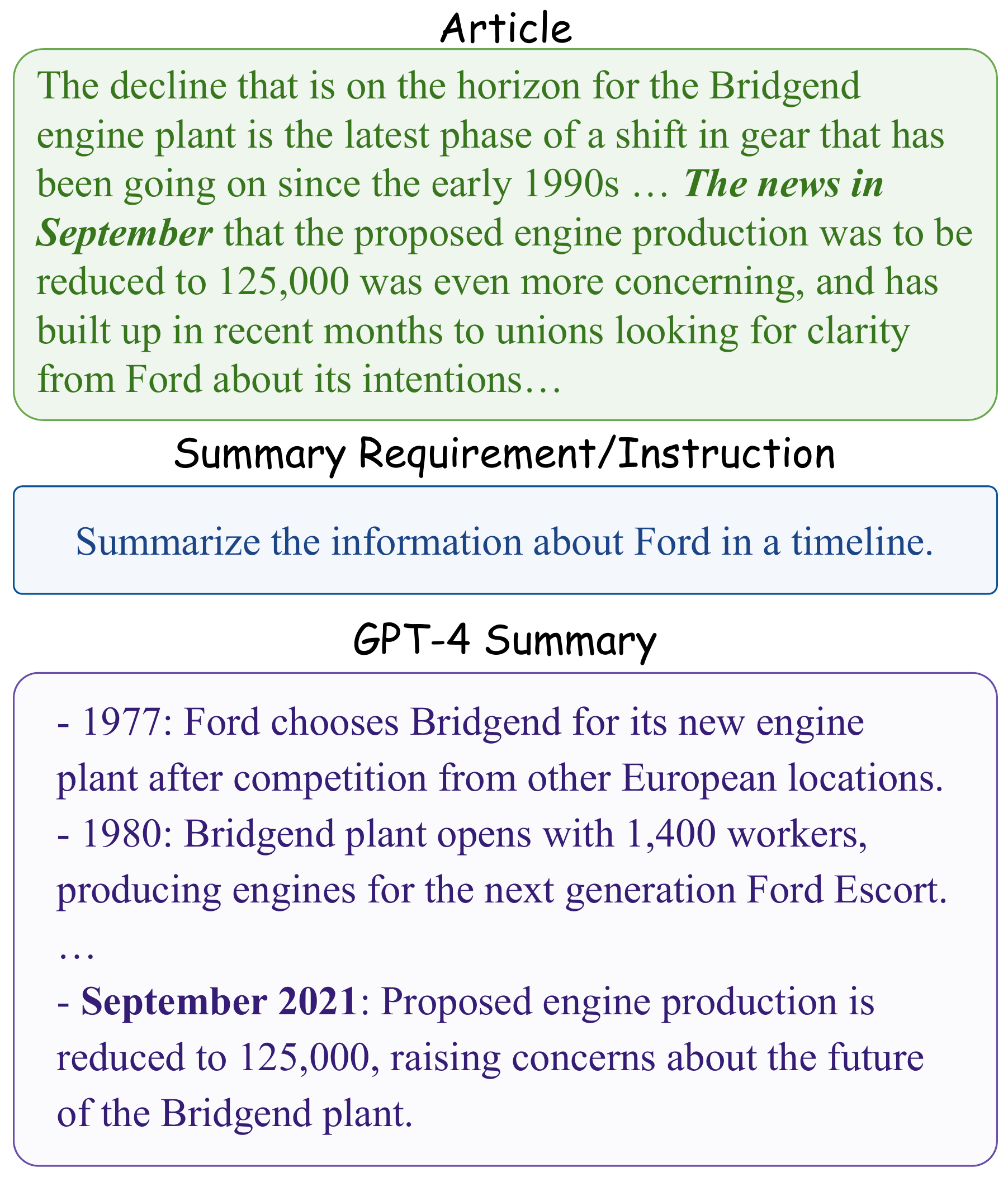}
 \caption{\label{fig:example}Task example of instruction controllable text summarization. 
 GPT-4 made a factual error in the summarized timeline by including a fabricated date (2021) not mentioned in the source article.
}
\end{figure}

Recent studies~\citep{goyal2022news, liu-etal-2023-revisiting, 10.1162/tacl_a_00632, pu2023summarization} have found that large language models (LLMs), e.g., GPT-3.5~\citep{NEURIPS2022_b1efde53}, can achieve state-of-the-art, even human-level performance on widely used summarization benchmarks such as the CNN/DailyMail dataset~\cite{nallapati-etal-2016-abstractive}.
Moreover, there are signs that LLM performance is saturated on the task of generic summarization, since on these benchmarks
(1) LLMs with varying capacity levels, e.g., GPT-3.5 and GPT-4~\citep{OpenAI2023GPT4TR}, are rated similar under human evaluation~\citep{10.1162/tacl_a_00632, pu2023summarization};
(2) the inter-annotator agreement for comparing strong-performing systems is usually low and significantly influenced by subjective preferences~\citep{goyal2022news, liu-etal-2023-revisiting, 10.1162/tacl_a_00632}.

We argue that a root cause of this saturation is that traditional summarization settings, such as ``\textit{summarize this article in a few sentences,}'' can be too simplistic and underconstrained \cite{kryscinski-etal-2019-neural}; without specifying the information need of an intended user, there exist many ``good'' summaries, but no clear criteria to compare them.
Consequently, the summaries generated under these settings may not fully satisfy practical usability criteria, and it remains an open question whether LLMs can perform well in more controlled settings aligned with users' real needs.

Therefore, we aim to study LLMs' capacities in \textit{\textbf{instruction controllable}}\footnote{For brevity, we will use the term ``ins-controllable'' to refer to ``instruction controllable'' throughout this paper.} text summarization.
To this end, we define a summarization task that takes both a source article and a summary requirement/instruction as input.
This task setting can be viewed as an extension of both query-focused summarization~\citep{zhong2021qmsum,vig-etal-2022-exploring} and aspect/attributed-based controllable summarization~\citep{he-etal-2022-ctrlsum, 10.1162/tacl_a_00575}.
However, the natural language instructions offer greater controllability and flexibility for more complex situations, such as a combination of an information query and a formatting requirement, leveraging the LLMs' instruction-following abilities~\citep{NEURIPS2022_b1efde53}.
We show a task example in Figure~\ref{fig:example}.

To study our proposed task setting, \textbf{we curate a human annotation benchmark that evaluates the performance of several representative LLMs on ins-controllable summarization} (\S\ref{sec:annotation}).
Specifically, we construct task samples by manually selecting articles from the XL-Sum dataset~\citep{hasan-etal-2021-xl} and writing the summary requirements ourselves, aiming to reflect the actual information needs of the users during reading.
Then, we collect human annotations of representative LLMs on this task along 4 quality dimensions: (1) overall quality, (2) missing information, (3) irrelevant information, and (4) factual consistency.
The evaluation results present a comprehensive view of the current LLMs performance on the ins-controllable summarization task, demonstrating large performance gaps among LLMs with different capacities.
Furthermore, we found that even the strongest LLM that we evaluated, i.e., GPT-4, still makes factual and other types of errors, indicating room for future improvement.

During the human annotation collection, we found that as the complexity of the summarization task rises, evaluating the summaries becomes increasingly difficult.
Therefore, \textbf{we investigate the performance of a variety of LLM-based automatic evaluation methods on our proposed task} (\S\ref{sec:meta-eval}).
To this end, we compare 40 evaluation methods, each a combination of an evaluation protocol, such as point-wise scoring (e.g., G-Eval~\citep{liu-etal-2023-g}), and an LLM as the backbone model.
Using the collected human annotations to evaluate these methods, we observe significant performance gaps among different evaluation protocols and different LLMs.
Moreover, we found that while several methods we investigate, such as the GPT-4 powered ones, already achieve a strong performance at comparing summarization systems, none of them are well-aligned with the human evaluation when comparing individual summaries.

Having identified the most reliable automatic evaluation methods, e.g., pairwise comparison powered by GPT-4, \textbf{we investigate whether these evaluation methods can reliably automate the benchmarking of ins-controllable summarization} (\S\ref{sec:benchmark}).
Specifically, we evaluated 11 different LLMs along the quality dimensions we defined.
We found that the current LLM-based evaluation methods fail to provide convincing results since they can be biased by confounding factors such as summary lengths.

Our contributions are as follows:

\noindent (1) We curated a manually annotated evaluation dataset for ins-controllable text summarization to facilitate the evaluation of LLM-based summarization and summarization evaluation. 

\noindent (2) We collected a human evaluation benchmark, \dataset, consisting of multi-dimensional quality annotations of summaries generated by different LLMs on the ins-controllable summarization task, and made \dataset publicly available.\footnote{\dataset is available at \url{https://github.com/yale-nlp/InstruSum}.}

\noindent (3) We benchmarked a series of LLM-based automatic evaluation methods that couple different evaluation protocols with various LLMs using our collected human annotations, and highlighted their limitations in automatic benchmarking for ins-controllable summarization.

\section{Human Annotation Collection}
\label{sec:annotation}

We curate a human evaluation benchmark for ins-controllable summarization with two steps:
(1) dataset creation and (2) system output evaluation.

\subsection{Dataset Creation}
\label{subsec:data-sample-collection}

\begin{table}
\small
\centering
    \extrarowheight=\aboverulesep
    \addtolength{\extrarowheight}{\belowrulesep}
    \aboverulesep=0pt
    \belowrulesep=0pt
    \begin{tabular}{p{0.94\linewidth}}
\toprule
Summarize the main factors that led to the conflict between the Ethiopian government and forces in the Tigray region. \\
\rowcolor{gray!10} Summarize the notable figures from the Prohibition era mentioned in this article. \\
Summarize the history of the Cononish gold mine in bullet points. \\
\rowcolor{gray!10} Summarize the views of Democrats and Republicans on trusting information coming from the WHO. \\
Summarize the experiences of Chum Mey in the 1970s with a timeline.  \\
\rowcolor{gray!10} Summarize the similarities of the definitions of collusion provided by different people in this article. \\
Summarize the concerns and opposition from the public about the new PNR directive into bullet points for the views of each group. \\
\rowcolor{gray!10} Summarize the possible explanations for why there hasn't been any firm evidence of aliens' existence, under the assumption that they do exist. \\
Summarize the people quoted in the article and their identity. \\
\rowcolor{gray!10}  Summarize the efforts of the Brazilian Tourist Board to attract more tourists in three sentences. \\
\bottomrule
\end{tabular} 
    \caption{\dataset summary requirement examples.}
    \label{tab:examples}
\end{table}

Ins-controllable summarization can be defined as
\begin{equation}
\scalebox{0.9}{
$S \displaystyle \leftarrow f(D, I)$,}
\end{equation}
where $D$ is the input document, $I$ is a specific summary requirement, $S$ is the desired summary, and $f$ is a summarization system.
To ensure the quality of our following evaluation, we (the authors) manually construct an evaluation-only dataset with the proposed task format.
The articles chosen are from the English split of XL-Sum~\citep{hasan-etal-2021-xl} dataset, containing news articles from the BBC website.
We chose XL-Sum because 
(1) XL-Sum was already made public by \citet{hasan-etal-2021-xl}, which makes it easier for us to release our benchmark;
(2) XL-Sum is newer and less commonly studied than other datasets such as CNN/DailyMail~\citep{nallapati-etal-2016-abstractive}, which reduces the concern of data contamination.
We collected 100 article-requirement pairs in total with the following steps:

\noindent (1) \textbf{Searching for challenging articles}.
Since not all XL-Sum articles are complicated enough to require specialized summaries, we first select articles with abundant and complex information, of which the requirement-specific summaries can be very beneficial to the readers.
Besides, only articles with around 1000-1200 words are selected to ensure they are sufficiently long but not too difficult for human evaluation.  

\noindent (2) \textbf{Writing summary requirements}.
After selecting an article, we write one or more summary requirements with different focuses, simulating the real reading experience, where readers may have different informational needs throughout the reading process, as well as structural or formatting preferences.
We also used GPT-4 to generate candidate requirements in order to increase the requirement diversity.\footnote{The prompt template is in Appendix~\ref{subsec:appendix-prompts-collection}.} However, they were not frequently used and were edited by us to ensure their naturalness and correctness.
In Table~\ref{tab:examples}, we show a list of summary requirements.

\noindent (3) \textbf{Obtaining hybrid LLM-human summary}.
With the article-requirement pair, we prompt the LLMs to generate a summary using a zero-shot prompt.\footnote{The prompt template is in Appendix~\ref{subsec:appendix-prompts-collection}.}
We then make minimal necessary edits to the LLM summary to obtain a hybrid LLM-human summary.
To analyze the effect of the choice of LLM on the human-edited summary, we interchangeably used three OpenAI LLMs to generate the initial summary: \texttt{text-davinci-003}, \texttt{gpt-3.5-turbo-0301}, and \texttt{gpt-4-0314}.\footnote{\url{https://platform.openai.com/docs/models}} 

The basic dataset statistics are in Table~\ref{tab:dataset-statistics}.

\begin{table}
    \centering
    \small
\addtolength{\tabcolsep}{-2pt} 
    \begin{tabular}{cccc}
\toprule
   \textbf{Article} &   \textbf{Requirement} &   \textbf{Initial Summ.} &   \textbf{Hybrid Summ.} \\
\midrule
1193.4&15.4&115.1&107.7 \\
\bottomrule
\end{tabular}
\addtolength{\tabcolsep}{2pt} 
    \caption{Dataset statistics of \dataset. The average length (tokens) of the article, the summary requirement, the initial LLM summary, and the hybrid LLM-human summary are reported.}
    \label{tab:dataset-statistics}
\end{table}

\subsection{System Output Evaluation}
\label{subsec:sys-human-eval}
We benchmark the LLMs' performance on the ins-controllable summarization task by collecting human judgments over 4 quality dimensions on the 100 samples from above, resulting in a new benchmark, \dataset, consisting of 500 summary-level human annotations. The dimensions are:

\noindent (1) \textbf{Overall Quality}: This rating assesses the overall quality of the summary in relation to the summary requirement.

\noindent (2) \textbf{Missing Information}: Does the summary omit any crucial information from the article concerning the summary requirement?

\noindent (3) \textbf{Irrelevant Information}: Does the summary include any information that is not relevant to the summary requirement?

\noindent (4) \textbf{Factual Consistency}: Is the summary consistent with the facts presented in the article, without contradicting or misrepresenting any information?

We annotate each quality dimension using a ranking protocol,
ranking summaries from 1 (best) to 5 (worst).
For factual consistency, we ask the annotators to select the span(s) containing a factual inconsistency, and for overall preference, we ask the annotators to explain the reasoning behind their overall rankings. 
Screenshots of our annotation protocol can be found in Appendix~\ref{sec:appendix-interface}.
\par 
We select the following four LLMs for annotation: \texttt{text-davinci-002}, \texttt{text-davinci-003}, \texttt{gpt-3.5-turbo-0301}, \texttt{gpt-4-0314}, in addition to the hybrid LLM-human summary.
These models are chosen to help study recent LLM development over multiple sizes and training paradigms.\footnote{Model details are in Appendix~\ref{sec:appendix-model-index}.}
For each summary, we collect three annotations.

\begin{table}
    \centering
    \small
    \begin{tabular}{lrrrr}
\toprule
\textbf{Mode}   &   \textbf{Overall} &   \textbf{Missing} &   \textbf{Irrelevant} &   \textbf{Factual} \\
\midrule
 listwise   &     0.2571 &     0.2247 &        0.1925 &     0.0196 \\
 pairwise   &     0.4428 &     0.3657 &        0.2588 &      0.0721 \\
\bottomrule
\end{tabular}
    \caption{Inter-annotator agreements (Krippendorff's alpha) for \dataset across various quality dimensions at both \textit{listwise} ranking and \textit{pairwise} comparison levels.}
\vspace{-2mm}
    \label{tab:agreement}
\end{table}

For this annotation, we recruit annotators on Amazon Mechanical Turk\footnote{\url{https://www.mturk.com/}} (MTurk).
The annotators must pass two rounds of qualification tests.
Moreover, to ensure the annotation quality, we maintained ongoing conversations with the annotators to exchange feedback and address their questions.
Additionally, we conducted spot checks on each batch of annotations to maintain quality and provide feedback to the crowd annotators.

The inter-annotator agreements are presented in Table~\ref{tab:agreement} for both the original ranking annotation task and the converted pairwise comparison results with the MASI distance~\citep{passonneau-2006-measuring} following  \citet{goyal2022news} to enhance comparability.
We are able to achieve moderately high agreements on most dimensions, including the overall quality evaluation, which has been shown to be difficult to annotate with high agreements in prior work \cite{10.1162/tacl_a_00632}. 
Regarding factual consistency, we note that low agreement may stem from the sparsity of errors in the dataset. 
We (the authors) manually verified whether the annotated spans contained factual errors.
Our annotation revealed that the errors made often proved to be subtle errors or nuanced different understandings of the article, and we found the accuracy of the crowd annotations to be 88.4\%.
Factual error examples are provided in Appendix~\ref{sec:appendix-fact-error}.

We note the difficulty of our annotation task.
Specifically,
(1) Earlier iterations of our annotation interface used a Likert scale, but we found that this resulted in a low inter-annotator agreement;
(2) Only around 5\% of the crowd-workers that participated in the qualification tests achieved acceptable performance to be recruited for our task, although all of them have at least a 99\% acceptance rate on MTurk;
(3) The average time to complete one annotation task is around 30 minutes. 
Furthermore, we increased the payment level to enhance annotator retention, as the high cognitive demands of our task tend to discourage annotators from completing more tasks. 
As a result, we found it challenging to expand the annotation sample size because of both budget constraints and the intensive labor required.

\subsection{Are LLMs Good at Ins-Controllable
Summarization?}
\label{subsec:human-eval-results}

\begin{table}
    \centering
    \small
\addtolength{\tabcolsep}{-4pt} 
    \begin{tabular}{lrrrr}
\toprule
\textbf{System}   &   \textbf{Overall} &   \textbf{Missing} &   \textbf{Irrelevant} &   \textbf{Factual} \\
\midrule
 text-davinci-002   &     2.344 &     2.595 &        3.443 &     0.640 \\
 text-davinci-003   &     3.239 &     3.703 &        3.708 &     0.710 \\
 gpt-3.5-turbo-0301 &     2.897 &     3.473 &        2.958 &     0.800 \\
 gpt-4-0314         &     \textbf{3.970} &     \textbf{4.067} &        4.205 &     \textbf{0.860} \\
 hybrid             &     3.873 &     3.947 &        \textbf{4.359} &     \textbf{0.860} \\
\bottomrule
\end{tabular}
\addtolength{\tabcolsep}{4pt} 
    \caption{\dataset: human evaluation results of LLM-generated ins-controllable summaries on 4 quality dimensions.
    The scores for \textit{overall} quality, \textit{missing} information, and \textit{irrelevant} information dimensions range from 1 to 5.
    The \textit{factual} score is the ratio of factually consistent summaries.
    \textit{Hybrid} is the hybrid LLM-human summary.}
\vspace{-2mm}
    \label{tab:human-eval}
\end{table}

Results from our human evaluation are found in Table \ref{tab:human-eval}.
For the \textit{overall quality}, \textit{irrelevant information}, \textit{missing information} dimensions, we convert the human-annotated rankings we obtained in \S\ref{subsec:sys-human-eval} into system scores as follows on each data example:
\begin{equation}
    \scalebox{0.9}{$\displaystyle s_i = N - \sum_{j=1}^{N} \mathbbm{1}_{\{r < r_i\}}(r_j),$}
\end{equation}
where $s_i$ is the converted score of the summary of the $i$-th system, $r_i$, $r_j$ are the summary rankings and a smaller ranking represents higher quality, $\mathbbm{1}$ is the indicator function, and $N$ is the number of ranked summaries.
Using this scoring schema, a perfect system would achieve a full score of $N$ (i.e., 5 in our case).
For factual consistency, since we found a low agreement in the crowd annotations, we report the ratio of factual summaries according to the human annotations verified by the authors.
We note the following findings from Table \ref{tab:human-eval}:

\noindent (1) \textbf{There is a large performance gap among the different LLMs}.
Specifically, GPT-4 is significantly better than the GPT-3.5 models,\footnote{The p-value is less than 0.01 for all the comparisons, except for \texttt{gpt-4-0314 } v.s. \texttt{gpt-3.5-turbo-0301} on the factual consistency dimension, of which the p-value is 0.058. }
and the supervisedly fine-tuned \texttt{text-davinci-002} archives worse performance than the LLMs fine-tuned with reinforcement learning from human feedback.\footnote{Model details are in Appendix~\ref{sec:appendix-model-index}.}
In contrast, recent work~\citep{pu2023summarization} found that the performance of GPT-3.5 and GPT-4 are very similar on the generic news summarization task.
This suggests that the ins-controllable summarization task we proposed can be a more suitable benchmarking task for the LLMs.

\noindent (2) \textbf{All LLMs we evaluated still make a considerable amount of factual errors in their summaries}.
For example, the error rates of \texttt{text-davinci-002} and \texttt{gpt-4-0314} are 36\% and 14\%, respectively.
This is also different from the patterns on the generic news summarization task, on which the factual error rate is only 1-2\% for \texttt{text-davinci-002}~\citep{10.1162/tacl_a_00632}.

\noindent (3) \textbf{The hybrid LLM-human summary can only outperform GPT-4 on the irrelevant information dimension}, suggesting that GPT-4 is close to the human-level performance, especially when the human annotator is asked to only edit the LLM summary.
It also indicates that generating summaries without irrelevant information is the most challenging dimension for current LLMs.
Interestingly, the following section (\S\ref{subsec:meta-eval-result}) will show that irrelevant information is also the most challenging dimension for the LLM-based automatic evaluation methods.
 We provide a fine-grained comparison of the initial LLM and hybrid summaries in Appendix~\ref{sec:appendix-human-eval-results}.\footnote{We interchangeably used GPT-4 and GPT-3.5 models to generate the initial LLM summary so Table \ref{tab:human-eval} does not provide a direct comparison between the initial and hybrid summaries.}

\begin{table}[t]
    \centering
    \small
   \begin{tabular}{llc}
\toprule
 \textbf{Dimension1}    & \textbf{Dimension2}    &   \textbf{Agreement} \\
\midrule
 Overall    & Irrelevant &       0.412 \\
 Overall    & Missing    &       \textbf{0.611} \\
 Irrelevant & Missing    &       0.209 \\
\bottomrule
\end{tabular}
    \caption{The inter-dimension agreements of human annotations on \dataset. \textit{Overall quality} and \textit{missing information} dimensions have the highest agreements.}
    \label{tab:inter-dimension-human}
\end{table}

\paragraph{Inter-Dimension Analysis}
To explore the relationship between human evaluation results across different quality dimensions, we examine the (listwise ranking) agreements between the summary scores for these dimensions in Table \ref{tab:inter-dimension-human}.
The results show that the \textit{missing information} dimension has a higher influence on the \textit{overall quality} dimension than the \textit{irrelevant information} dimension, suggesting that human annotators favor comprehensive over concise summaries, similar to recent work's findings on length bias~\citep{liu-etal-2023-revisiting, singhal2023long,zheng2023judging,huang2023embrace,saito2023verbosity}
in human evaluation.

\section{Are LLMs Good at Ins-Controllable Summary Evaluation?}
\label{sec:meta-eval}

Human evaluation of our proposed ins-controllable summarization is complex and time-intensive, requiring scalable, reliable automatic evaluation methods. Consequently, we benchmark recent LLM-based evaluation methods, exploring various evaluation protocols and LLM backbones.

\subsection{LLM-based Evaluation Protocols}

LLM-based automatic evaluation methods can be categorized along two orthogonal dimensions -- the \textbf{\textit{evaluation protocols}} and the backbone LLMs.
We investigate the following evaluation protocols:

\noindent (1) \textbf{LLMScore}: direct scoring using predicted probability.
GPTScore~\citep{fu2023gptscore} proposes a protocol that interprets the LLM-predicted probability of certain token(s) as a quality score.

\noindent (2) \textbf{LLMEval}: direct scoring by text completion.
In \citet{chiang-lee-2023-large} and G-Eval~\citep{liu-etal-2023-g}, the LLM is asked to assign a quality score.

\noindent (3) \textbf{LLMCompare}: pairwise comparison between two candidate outputs by text completion~\citep{zheng2023judging, wang2024pandalm}.

\noindent (4) \textbf{LLMRank}: listwise ranking by text completion, simultaneously evaluating a list of candidate outputs~\citep{sun-etal-2023-chatgpt, liu2023learning}.

\noindent \textbf{Prompt Design}.
For each of the evaluation protocols, we design \textit{dimension-specific} prompts for the evaluation quality dimensions we defined in \S\ref{subsec:sys-human-eval}.
The prompt templates are in Appendix~\ref{subsec:appendix-prompt-llmeval}.

\begin{table*}[ht]
    \centering
\small
\addtolength{\tabcolsep}{-5.2pt} 
    \begin{tabular}{@{\extracolsep{1pt}}lrrrrrrrr@{}}
\toprule
& \multicolumn{4}{c}{\textbf{System-level Correlations}} & \multicolumn{4}{c}{\textbf{Summary-level Correlations}} \\
\cmidrule{2-5} \cmidrule{6-9}
              &   \textbf{LLMRank} &   \textbf{LLMCompare} &   \textbf{LLMEval} & \textbf{LLMScore}  & \textbf{LLMRank} &   \textbf{LLMCompare} &   \textbf{LLMEval} & \textbf{LLMScore}   \\
\cmidrule{2-5} \cmidrule{6-9}
& \multicolumn{8}{c}{\textbf{Overall Quality}} \\
\cmidrule{2-9}
gpt-3.5-turbo-0301     &     0.738 &        0.400 &     0.600 & -          &     0.005 &        0.185 &     0.223 & -          \\
 gpt-3.5-turbo-0613     &     0.600 &        0.527 &     0.527 & -          &    -0.012 &        0.160 &     0.048 & -          \\
 gpt-4-0314             &     0.800 &        \textbf{1.000} &     \textbf{1.000} & -          &     0.095 &        0.361 &     0.271 & -          \\
 gpt-4-1106-preview     &     0.400 &        0.800 &     0.800 & -          &     0.047 &        \textbf{0.483} &     0.257 & -          \\
 text-davinci-002       &    -0.200 &        0.400 &     0.738 & 0.600      &    -0.044 &        0.026 &     0.114 & 0.062      \\
 text-davinci-003       &     0.400 &        0.400 &     0.949 & -0.400     &    -0.034 &        0.029 &     0.052 & -0.133     \\
 gpt-3.5-turbo-instruct &     0.400 &        0.600 &     0.738 & -0.200     &     0.006 &        0.212 &     0.078 & -0.058     \\
 llama-2-7b-chat             &     0.200 &        0.527 &     0.527 & 0.000      &    -0.062 &       -0.019 &     0.028 & 0.063      \\
 llama-2-13b-chat            &     0.105 &        0.400 &     \textbf{1.000} & -0.400     &    -0.058 &        0.096 &     0.037 & -0.032     \\
 llama-2-70b-chat            &    -0.316 &        0.400 &     0.949 & 0.800      &    -0.006 &        0.072 &     0.016 & 0.116      \\
 mistral-instruct                  &    -0.400 &        0.200 &     0.447 & -0.200     &    -0.074 &        0.139 &     0.021 & 0.137        \\
 \cmidrule{2-9}
 & \multicolumn{8}{c}{\textbf{Missing Information}} \\
\cmidrule{2-9}
 gpt-3.5-turbo-0301     &     0.400 &        0.400 &     0.600 & -          &    -0.051 &        0.283 &     0.175 & -          \\
 gpt-3.5-turbo-0613     &     0.316 &        0.200 &     0.400 & -          &    -0.083 &        0.244 &     0.200 & -          \\
 gpt-4-0314             &     0.949 &        \textbf{1.000} &     0.949 & -          &    -0.001 &        0.440 &     0.233 & -          \\
 gpt-4-1106-preview     &     0.738 &        0.400 &     \textbf{1.000} & -          &     0.063 &        \textbf{0.443} &     0.085 & -          \\
 text-davinci-002       &     0.200 &        0.200 &     0.200 & 0.800      &    -0.034 &        0.037 &    -0.001 & 0.259      \\
 text-davinci-003       &     0.400 &        0.400 &     \textbf{1.000} & 0.400      &    -0.077 &        0.141 &     0.106 & 0.190      \\
 gpt-3.5-turbo-instruct &     0.200 &        0.600 &     0.738 & 0.800      &    -0.038 &        0.226 &     0.129 & 0.140      \\
 llama-2-7b-chat             &    -0.400 &        0.738 &     0.105 & -0.200     &    -0.108 &        0.012 &     0.016 & -0.103     \\
 llama-2-13b-chat            &     0.527 &        0.400 &     0.600 & 0.000      &    -0.051 &        0.246 &     0.085 & -0.046     \\
 llama-2-70b-chat            &     0.527 &        0.400 &     0.600 & -0.600     &    -0.023 &        0.119 &     0.044 & -0.173     \\
 mistral-instruct                &    -0.600 &        0.600 &     0.400 & 0.000      &    -0.120 &        0.205 &     0.061 & 0.036          \\
  \cmidrule{2-9}
 & \multicolumn{8}{c}{\textbf{Irrelevant Information}} \\
\cmidrule{2-9}
gpt-3.5-turbo-0301     &    -0.200 &       -0.200 &     0.200 & -          &    -0.008 &       -0.081 &     0.013 & -          \\
 gpt-3.5-turbo-0613     &     0.000 &        0.000 &    -0.200 & -          &    -0.007 &       -0.024 &    -0.026 & -          \\
 gpt-4-0314             &     0.400 &        0.600 &    \textbf{0.738} & -          &     0.057 &        0.208 &     0.057 & -          \\
 gpt-4-1106-preview     &     0.200 &        0.600 &     0.600 & -          &     0.180 &        \textbf{0.332} &     0.242 & -          \\
 text-davinci-002       &    -0.400 &       -0.400 &     0.105 & 0.200      &    -0.043 &       -0.053 &     0.067 & -0.062     \\
 text-davinci-003       &     0.000 &        0.105 &     0.600 & -0.400     &    -0.019 &       -0.009 &     0.139 & 0.058      \\
 gpt-3.5-turbo-instruct &     0.200 &        0.200 &     0.120 & -0.200     &     0.023 &        0.006 &     0.118 & 0.013      \\
 llama-2-7b-chat             &     0.000 &        0.200 &     0.000 & -0.600     &    -0.010 &        0.037 &    -0.029 & -0.064     \\
 llama-2-13b-chat            &     0.600 &        0.000 &     0.400 & 0.200      &    -0.012 &       -0.102 &    -0.004 & -0.011     \\
 llama-2-70b-chat            &    -0.105 &       -0.200 &     0.400 & -0.800     &    -0.042 &       -0.035 &     0.062 & 0.130      \\
 mistral-instruct          &    -0.527 &        0.000 &     0.200 & -0.200     &    -0.052 &       -0.095 &     0.046 & -0.095     \\
\bottomrule
\end{tabular}
\addtolength{\tabcolsep}{5.2pt} 
    \caption{Kendall rank correlations at both the system and summary levels between human evaluation and LLM-based evaluation over three quality dimensions on \dataset. 
    The LLM-based evaluation performance of different combinations of backbone LLMs (e.g., \texttt{gpt-4-0314}) and evaluation protocols (e.g., LLMRank) is reported.
    The best performance on each quality dimension at the system level or the summary level is highlighted. 
    }
    \label{tab:meta-eval}
\end{table*}

\subsection{Evaluation Settings}
\label{subsec:metaeval-setting}

We benchmark 11 LLMs in total on ins-controllable summarization evaluation over three quality dimensions, \textit{overall quality}, \textit{missing information}, and \textit{irrelevant information}.
We did not benchmark factual consistency evaluation since it is a more unique dimension, which we leave for more dedicated future work. 
For proprietary LLMs, we use different versions of GPT-3.5 and GPT-4 models provided by OpanAI.\footnote{\url{https://platform.openai.com/docs/models}}
For open-source LLMs, we benchmark LLama-2-chat~\citep{touvron2023llama} 7B, 13B, and 70B models, and the Mistral-Instruct~\citep{jiang2023mistral} 7B model.\footnote{Llama-2 models were released in July 2023, and Mistral 7B was released in September 2023.}
The full model list is in Table~\ref{tab:meta-eval}.

To compare LLM evaluation against human evaluation, we calculate the correlations between their evaluation scores at both system and summary levels.
The \textit{system-level} correlations measure how good the LLMs are at comparing summarization system performance, while the \textit{summary-level} correlations measure how good the LLMs are at comparing summary quality on individual data samples.
Since we adopted a ranking-based evaluation protocol for our human evaluation collection (\S\ref{subsec:sys-human-eval}), we use the Kendall rank correlation coefficient as the correlation measurement, and we transform the evaluation results of different evaluation protocols into rankings.
More evaluation setting details including the formal definitions of the correlation measurement are in Appendix~\ref{sec:appendix-meta-eval}.

\subsection{Result Analysis}
\label{subsec:meta-eval-result}

\begin{figure}[t!]
    \centering
    \includegraphics[width=1.0\linewidth]{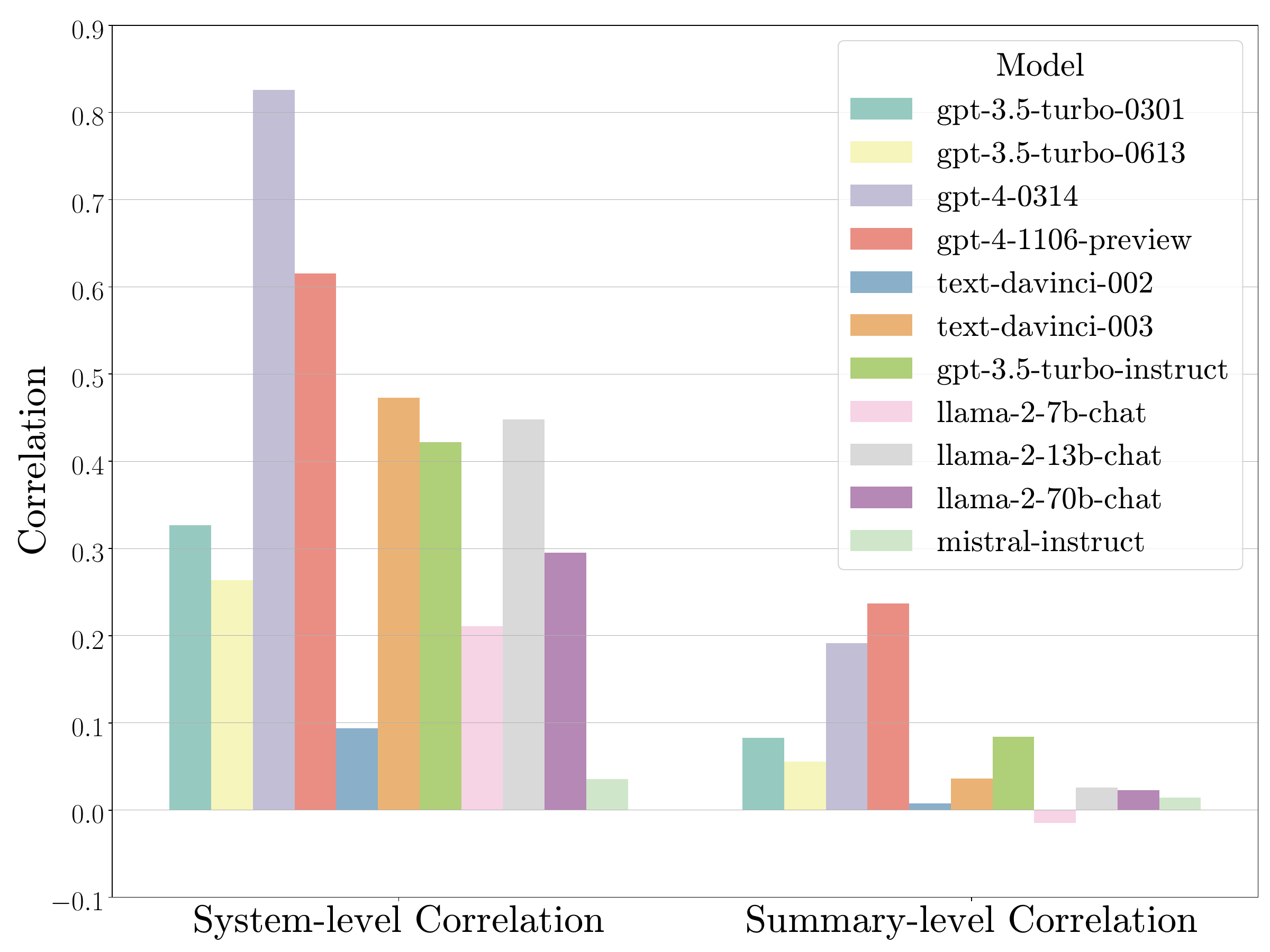}
 \caption{\label{fig:corrs}Average LLM performance of ins-controllable summary evaluation across 3 quality dimensions with 3 evaluation protocols on \dataset.
    }
    \vspace{-3mm}
\end{figure}

The evaluation results are reported in Table \ref{tab:meta-eval}.
For each of the backbone LLMs we chose to benchmark (\S\ref{subsec:metaeval-setting}), we evaluate its performance with different evaluation protocols when applicable\footnote{We could not use a few OpenAI models with LLMScore since the log-likelihood is not provided by the APIs.} so in total we evaluate 40 LLM-based evaluation methods.
We make the following observations:

\noindent (1) \textbf{Different LLMs have significantly different performance at evaluating ins-controllable summaries}.
In Figure~\ref{fig:corrs}, we report the average LLM performance across 3 quality dimensions on 3 protocols except LLMScore.
In particular, GPT-4 shows a consistent advantage over other LLMs.

\noindent (2) \textbf{The choice of evaluation protocols has a large impact on the evaluation method performance}.
For example, the pairwise comparison protocol (LLMCompare) is (almost) always better than the listwise protocol (LLMRank).
Besides, the most suitable protocol for each backbone LLM can be different.
For instance, \texttt{gpt-4-0314} works better with LLMCompare while \texttt{gpt-3.5-turbo-0301} tends to work better with LLMEval.

\noindent (3) In general, LLM-based evaluation methods have much higher system-level correlations than summary-level correlations, which means these methods are better at evaluating which system is better on average, but \textbf{struggle at ranking different summaries of individual data examples}.
Notably, the strongest evaluation method we identified, i.e., pairwise comparison using \texttt{gpt-4-0314}, can only achieve an agreement value of 0.277 with human evaluation on the \textit{overall quality} dimension in pairwise comparison, lower than the human inter-annotator agreement (0.4428).

\noindent (4) \textbf{The performance of the LLM-based evaluation methods differs on different quality dimensions}.
In particular, \textit{Irrelevant Information} is a more challenging dimension than \textit{Missing Information}, suggesting that these methods are better at recall-based than precision-based evaluation.

\paragraph{Evaluation Consistency}
A reliable evaluator must yield consistent results across different evaluation protocols.
To check this consistency requirement, we examine the LLMs by calculating the correlations of its evaluation results on the LLMCompare and LLMEval protocols over the three quality dimensions, since these two protocols are most reliable.
Table~\ref{tab:eval-consistency} indicates low summary-level consistency among all evaluated LLMs. 
However, \texttt{gpt-4-0314} demonstrates a high system-level consistency, indicating it is the most reliable evaluator.

\begin{table}
    \centering
    \small
\addtolength{\tabcolsep}{-1.5pt} 
    \begin{tabular}{lrr}
\toprule
\textbf{System}   &   \textbf{System-Level} & \textbf{Summary-Level} \\
\midrule
gpt-3.5-turbo-0301     &          0.600 &           0.149 \\
 gpt-3.5-turbo-0613     &          0.681 &           0.135 \\
 gpt-4-0314             &          \textbf{0.966} &           0.227 \\
  gpt-4-1106-preview     &          0.800 &           \textbf{0.262} \\
 text-davinci-002       &          0.418 &           0.049 \\
 text-davinci-003       &          0.485 &           0.089 \\
 gpt-3.5-turbo-instruct &          0.461 &           0.114 \\
 llama-2-7b-chat             &          0.111 &          -0.006 \\
 llama-2-13b-chat            &          0.200 &           0.072 \\
 llama-2-70b-chat            &          0.442 &           0.021 \\
 mistral-instruct                &            0.416 &           0.051 \\ 
\bottomrule
\end{tabular}
\vspace{-2mm}
\addtolength{\tabcolsep}{1.5pt} 
    \caption{LLM evaluation consistency between the LLMCompare and LLMEval protocols. System-level and summary-level Kendall rank correlations are reported.}
    \label{tab:eval-consistency}
\vspace{-2mm}
\end{table}

\begin{table}
    \centering
    \small
\addtolength{\tabcolsep}{-4,0pt} 
    \begin{tabular}{lccc}
\toprule
 \multirow{2}{*}[-2pt]{\textbf{System}}         &   \multirow{2}{*}[-2pt]{\textbf{Generator}} &  \multicolumn{2}{c}{\textbf{Re-ranker (Evaluator)}} \\
 \cmidrule{3-4}
 & &\textbf{LLMCompare} &\textbf{LLMEval}  \\
\midrule
  text-davinci-002   &        2.344 &        3.335 &     3.383 \\
 text-davinci-003   &        3.239 &        3.189 &     3.374 \\
 gpt-3.5-turbo-0301 &        2.897 &        3.357 &     3.504 \\
 gpt-4-0314         &        3.970 &        3.533 &     3.561 \\
\bottomrule
\end{tabular}
\addtolength{\tabcolsep}{4.0pt} 
    \caption{Performance comparison of LLMs as the summary generator and the summary re-ranker with different evaluation protocols. The human-annotated scores on the overall quality dimension are reported. A random-reranking oracle can achieve a score of 3.260.}
    \label{tab:gen-eval}
\vspace{-3mm}
\end{table}

\paragraph{Generator-Evaluator Consistency}
Recent work has found that the behavior and performance of LLMs can differ when they are used as a generator or an evaluator on the same task~\citep{li2024benchmarking, west2024the}.
Thus, we analyze the performance consistency of LLMs on ins-controllable summary generation and evaluation.
To this end, we treat the LLM evaluator as a \textit{re-ranker} to make its performance more comparable to the LLM generator.
In Table~\ref{tab:gen-eval}, we report the generation and re-ranking performance of the 4 human-evaluated LLMs in \S\ref{subsec:sys-human-eval} on the overall quality dimension, where the re-ranker can select its output from the human-evaluated candidate summaries.
We note: 

\noindent (1) The generation performance does not always align with the evaluation performance.
For example, \texttt{text-davinci-003} has better generation performance than \texttt{gpt-3.5-turbo-0301}, but worse evaluation performance.

\noindent (2) 
Despite its promising ability of summary generation, \texttt{gpt-4-0314} fails to outperform its generation performance under the re-ranking task setting, suggesting a lack of deeper task understanding.

\section{Can We Automate Ins-Controllable Summarization Benchmarking?}
\label{sec:benchmark}

\begin{figure}[t!]
    \centering
    \includegraphics[width=1.0\linewidth]{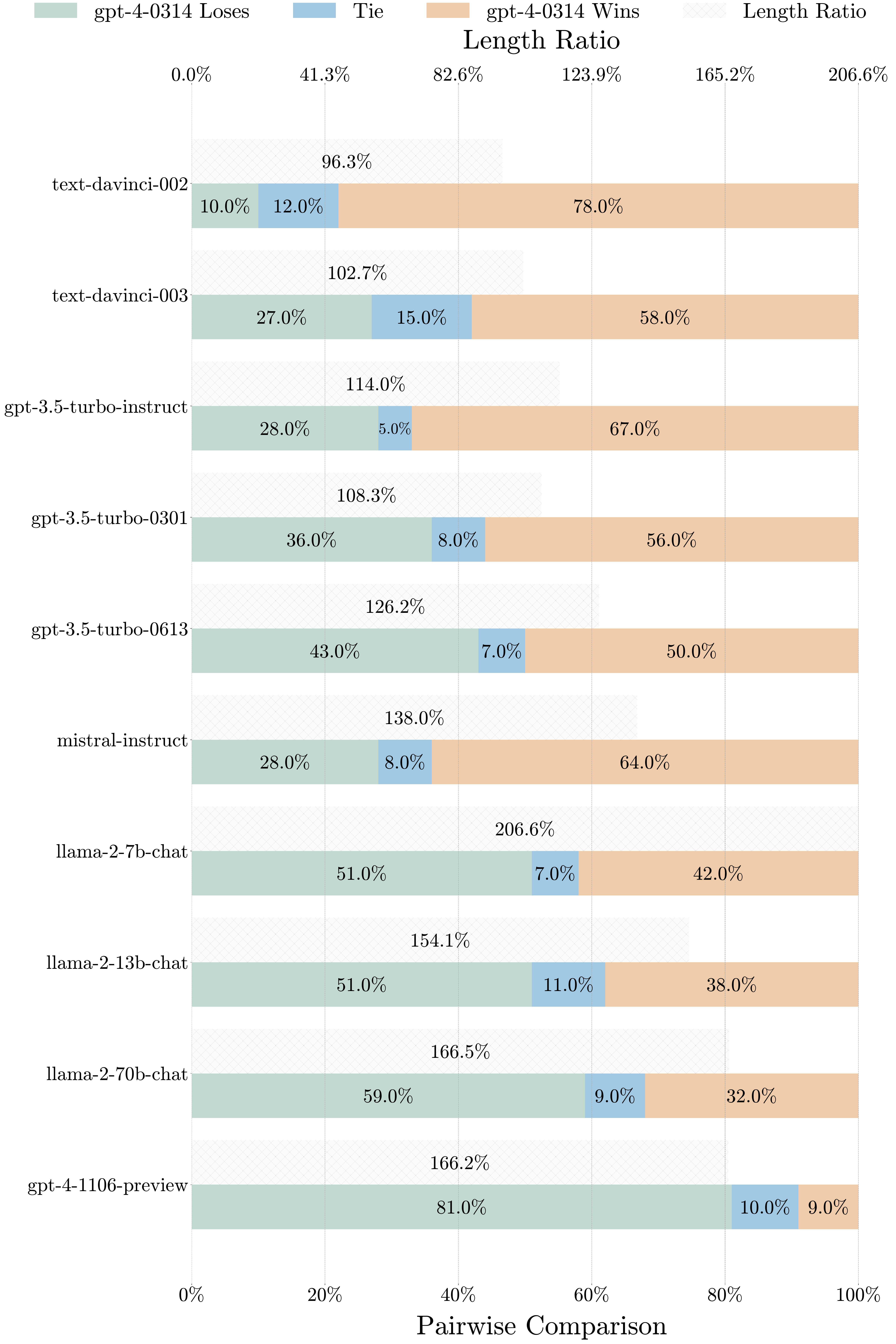}
\vspace{-7mm}
 \caption{\label{fig:autoeval}Automatic benchmarking results using \texttt{gpt-4-0314} as the evaluator.
 LLM summaries are compared against \texttt{gpt-4-0314}'s summaries. 
 The model evaluation result (bottom) is reported, as well as the summary length ratios (top) relative to \texttt{gpt-4-0314}.}
 \vspace{-3mm}
\end{figure}

After evaluating LLMs as summary evaluators, we explore their potential for automating ins-controllable summarization benchmarking.

\subsection{Evaluation Settings}
Since GPT-4 coupled with LLMCompare is the best evaluation method we identified in \S\ref{subsec:meta-eval-result}, we use it for the automatic benchmarking.
To avoid the prohibitive cost, we treat GPT-4 (\texttt{gpt-4-0314}) as a baseline and evaluate the other systems by comparing them against GPT-4 only, following recent practices in automatic LLM benchmarking~\citep{dubois2023alpacafarm, zheng2023judging}.%
\footnote{We randomly shuffled the summary pairs and did not observe a significant positional bias in the evaluation results.}
We evaluated 11 LLMs over the 100 data examples we used in human evaluation (\S\ref{subsec:data-sample-collection}).
The prompt template for summary generation is shown in Appendix~\ref{subsec:appendix-prompts-collection}.

\subsection{Result Analysis}
\label{subsec:benchmark-result}

\begin{table}
    \centering
    \small
    \begin{tabular}{lrrr}
\toprule
  \textbf{Pair} &   \textbf{Overall} &   \textbf{Missing} &   \textbf{Irrelevant}  \\
\midrule
 Human v.s. Oracle & 0.112 & -0.268 & 0.302 \\
 GPT-4 v.s. Oracle & 0.304 & 0.098 & 0.253 \\
 GPT-4 v.s. Human & 0.277 & 0.147 & 0.376 \\ 
\bottomrule
\end{tabular}
    \caption{Agreements among human evaluation, LLM-based evaluation (\texttt{gpt-4-0314}) and the length oracle.}
    \label{tab:length-orcale}
\end{table}

The evaluation results are in Figure~\ref{fig:autoeval}.
We found that Llama-2 models show a strong performance under GPT-4's evaluation, even outperforming \texttt{gpt-4-0314} in the pairwise comparison setting.
However, since we did not observe Llama-2 models achieving performance as strong as GPT-4 in evaluating the ins-controllable summaries (\S\ref{subsec:meta-eval-result}), we suspect GPT-4 has overestimated Llama-2 models' performance in this summary generation task.
The reason is likely that the summaries generated by Llama-2 models are much longer than the \texttt{gpt-4-0314} summaries, which tend to be favored by GPT-4 as shown below.

In Table~\ref{tab:length-orcale}, we compare the annotator agreement in pairwise comparison among human evaluation, LLM-based evaluation using \texttt{gpt-4-0314} and LLMCompare, and a length oracle that always prefers longer summaries.
The results indicate that human evaluation has a positive correlation with the length oracle in the \textit{irrelevant information} dimension and a negative correlation in the \textit{irrelevant information} dimension.
Conversely, \texttt{gpt-4-0314} has a positive correlation with the length oracle across all quality dimensions.
These findings suggest that LLM-based evaluation is more prone to bias from summary length compared to human evaluation.
Furthermore, our case study in Appendix~\ref{sec:appendix-bias} shows that when the length difference is controlled, none of the LLMs we evaluated have a clear advantage over \texttt{gpt-4-0314}.
Therefore, we find current LLM-based evaluation methods unreliable for automatic ins-controllable summarization benchmarking. 
To facilitate future studies in this direction, we collect a large-scale learderboard of automatic evaluators with 41 LLMs, \textsc{InsturSumEval}, which is available at \url{https://huggingface.co/spaces/yale-nlp/InstruSumEval}.\footnote{Detailed information is in Appendix~\ref{appendix:instrusumeval}.}

\section{Related Work}

\paragraph{Summarization Benchmarks}
Recent work in summarization benchmarks has focused on aggregating model outputs and annotating them according to specific quality dimensions~\cite{huang-etal-2020-achieved,bhandari-etal-2020-evaluating,SummHumanFeedback,zhang-bansal-2021-finding,SummEval,gao-wan-2022-dialsummeval}.
In the context of LLMs, \citet{laban2023llms} incorporated LLMs into the benchmark-construction process while \citet{maynez-etal-2023-benchmarking} benchmarked LLMs on conditional text generation tasks including summarization. 
A few recent studies~\cite{goyal2022news, liu-etal-2023-revisiting, 10.1162/tacl_a_00632, pu2023summarization} point to the strength of LLMs with respect to human-written (reference) summaries on generic news summarization. 
In this work, we present a benchmark task that poses challenges for current LLMs and allows for further development and model comparison.

\paragraph{Instruction-Following Evaluation}
\citet{NEURIPS2022_b1efde53} introduce InstructGPT, which learns to follow instructions by aligning to human preference feedback and builds on earlier alignment work in summarization \cite{SummHumanFeedback}. 
Following \citet{NEURIPS2022_b1efde53}, a line of work~\cite{wang-etal-2023-self-instruct, pmlr-v202-zhou23g, zeng2024evaluating} has investigated methods of improving and benchmarking the instruction-following capabilities of LLMs.
Regarding text summarization, instruction-following text summarization expands upon work in query-focused summarization \cite{zhong2021qmsum,vig-etal-2022-exploring,yang2023exploring,pagnoni-etal-2023-socratic}, aspect-based summarization~\citep{10.1162/tacl_a_00632, pu2023summarization, yang-etal-2023-oasum}, and controllable summarization more broadly \cite{dou-etal-2021-gsum,he-etal-2022-ctrlsum, 10.1162/tacl_a_00575, bao-etal-2023-gemini, ribeiro-etal-2023-generating, ravaut2023promptsum,adams-etal-2023-sparse,adams-etal-2023-generating,narayan2023conditional,pagnoni-etal-2023-socratic,pu-demberg-2023-chatgpt}. 
Closely related to work on query-focused summarization is the task of long-form question answering \cite{fan2019eli5}, and recent work has benchmarked current models and metrics with a focus on completeness and factuality \cite{xu-etal-2023-critical}.
In this work, we explore controllability in the context of instructions.
Compared with query-focused summarization, our task format allows for more complex use cases where the information queries can be combined with other user request categories such as the output format.
\citet{wang-etal-2023-instructive} extends query-focused summarization and curates an instructive dialog summarization dataset.
The most relevant work to our study is \citet{skopek-etal-2023-towards}, which develops a dataset consisting of human annotations on instruction-summary pairs. 
However, their evaluation focuses only on the instruction-following capacities, while our human evaluation is multi-dimensional and puts more focus on the models' text summarization capabilities.

\paragraph{LLM-based Automatic Evaluation and Its Meta-Evaluation}
A series of recent work has investigated leveraging LLMs for automatic evaluation~\citep{fu2023gptscore, chiang-lee-2023-large, liu-etal-2023-g,zheng2023judging, wang2024pandalm,sun-etal-2023-chatgpt,gao2023human,wang-etal-2023-chatgpt,li2024generative}.
While these studies have demonstrated LLMs' promising performance on various evaluation tasks such as summarization evaluation, other recent work has highlighted the limitations of LLM-based automatic evaluation methods.
Specifically, LLMs can have various biases in their evaluation results~\citep{koo2023benchmarking, wang2023large}, and they fail to align with human evaluation when evaluating close-performing systems~\citep{shen-etal-2023-large} or adversarial examples~\citep{zeng2024evaluating}. 
Our work provides a thorough meta-evaluation of LLM-based methods, focusing on diverse protocols and backbone LLMs for ins-controllable summarization evaluation.

\section{Conclusions}

In this work, we benchmarked large language models for instruction controllable summary generation and evaluation, and presented a new benchmark dataset, \dataset.
We found that several LLMs have already shown promising performance in generating ins-controllable summaries.
However, they lack robust holistic capabilities for this task since they still make a considerable amount of errors in their summaries and they can not reliability evaluate different candidate summaries.
Furthermore, we notice large gaps between the performance of different generations of LLMs on both ins-controllable summary generation and evaluation.
As we believe our proposed ins-controllable summarization setting is more realistic and can provide better usability, we call for future work along this direction to make the text summarization systems more beneficial to the actual users.

\section{Limitations}
Our analysis is limited to 100 examples for which we collected human annotations of ins-controllable summaries generated by different LLMs.
While more statistically significant conclusions could be drawn from a larger evaluation set, as noted above a much larger time and budget allocation would be required, and we encourage the community to apply our protocol to expand our evaluation set.

Due to sparsity and subtleties of factuality errors generated by current LLMs on our benchmark, we did not perform a meta-evaluation of LLMs as factuality evaluators, since it would require a larger collection to observe significant error patterns.
We leave a larger evaluation of the factual consistency of current models and error types for future work.
%


\section*{Acknowledgements}
We express our gratitude to Tanya Goyal for her insightful suggestions, and we thank the anonymous reviewers for their constructive comments.
We are grateful for the compute support provided by the Google TRC program.

\bibliography{anthology,custom}

\appendix

\section{Prompt Templates}
\label{sec:appendix-prompts}
Here we provide the prompt templates we used throughout this work.

\subsection{Prompts in Human Annotation Collection}
\label{subsec:appendix-prompts-collection}

\paragraph{Prompt for Summary Requirement Recommendation}
In \S\ref{subsec:data-sample-collection}, we used GPT-4 to generate candidate summary requirements to help the human annotation.
The prompt template is as follows:

\begin{quote}
Please generate a list of specific summary requirements for a given article.

Here are some requirement examples based on different articles:
1. Summarize the possible explanations for why there hasn't been any firm evidence of aliens' existence, under the assumption that they do exist.
2. Summarize the experience of Chum Mey in 1970s with a timeline.
3. Summarize why Shanghai and Hong Kong seem to outperform Beijing in education. 
4. Summarize all people and their identities in the article.
5. Summarize the negative outcomes of the lockdown.
6. Summarize the conclusion of the fraud case.
7. Summarize the opinions of Ronan Barry in the article.
8. Summarize the events of Margaret's debit card fraud in a timeline.
9. Summarize the aftermath of sexual harassment on Meena in one sentence.
10. Summarize the difficulties faced by Uber and Lyft now.

Here's an article: 
\{\{article\}\}

Please generate a list of specific summary requirements for this article.
\end{quote}

\paragraph{Prompt for Generating Requirement-Specific Summaries}
We used the following template to prompt the LLMs to generate the requirement-specific summaries.
\begin{quote}
    Summarize the following article based on the specific requirement.

Article:
\{\{article\}\}

Requirement: \{\{requirement\}\}

Summary:
\end{quote}

\subsection{Prompts for LLM-based Automatic Evaluation}
\label{subsec:appendix-prompt-llmeval}

In \S\ref{sec:meta-eval}, we analyze the performance of different LLM-based automatic evaluation methods.
We designed prompt templates for each evaluation protocol and each evaluation dimension, and slightly fine-tuned templates for several LLMs to ensure that they are able to follow the instructions as much as they can.
To enhance the LLM evaluation performance, for LLMCompare and LLMRank, we design chain-of-thought~\citep{NEURIPS2022_9d560961} style prompts -- before the LLM gives the actual answer, it is prompted to first generate an explanation of the answer, mimicking the thinking process of human evaluators.
We show the following prompt templates for all the evaluation protocols on the \textit{overall quality} dimension, and all the templates can be found in our code release.

\noindent (1) \textbf{Prompt template for LLMRank}.

\begin{quote}
In this task, you will be provided with a news article, a specific summary requirement, and a list of summaries numbered as follows: 1. Summary 1, 2. Summary 2, and so on.

The summaries are crafted to meet a specific summary requirement. Note that there may be identical summaries within the list.

Your task is to evaluate and rank the summaries in ascending order of their overall quality concerning the summary requirement. First, you will explain your ranking, and then you will provide the ranking of each summary. The ranking should be a number between 1 and 5, where 1 is the best and 5 is the worst.

Note: In case of a tie, do not skip a rank. For example, if Summary 1 has ranking 1 and Summary 2 and 3 both have ranking 2, then Summary 4 should be assigned a ranking of 3, not 4.

Please refer to the example below for the format of your response.

Example Response:
Explanation: ``Your explanation of the ranking.''
Ranking: ``The ranking, e.g., 1, 2, 2, 3, 4.''

Here are the actual article, the summary requirement, and the summaries:

Article:
\begin{verbatim}
{{Article}}
\end{verbatim}

Summary Requirement:
\begin{verbatim}
{{Requirement}}
\end{verbatim}

Summaries:

1. Summary 1: 
\begin{verbatim}
{{Summary 1}}
\end{verbatim}

2. Summary 2: 
\begin{verbatim}
{{Summary 2}}
\end{verbatim}

3. Summary 3: 
\begin{verbatim}
{{Summary 3}}
\end{verbatim}

4. Summary 4: 
\begin{verbatim}
{{Summary 4}}
\end{verbatim}

5. Summary 5: 
\begin{verbatim}
{{Summary 5}}
\end{verbatim}

\end{quote}

\noindent (2) \textbf{Prompt template for LLMCompare}.

\begin{quote}
    In this task, you will be provided with a news article, a specific summary requirement, and two summaries.

The summaries are crafted to meet a specific summary requirement. Note that there may be identical summaries.

Your task is to compare the overall quality of these two summaries concerning the summary requirement and pick the one that is better (there can be a tie).
First you will give an explanation of your decision then you will provide your decision in the format of 1 or 2 or tie.

Please refer to the example below for the format of your response.

Example Response:

Explanation: ``Your explanation here''.

Decision: 1 or 2 or tie.

Here are the actual article, the summary requirement, and two summaries:

Article:
\begin{verbatim}
{{Article}}
\end{verbatim}

Summary Requirement:
\begin{verbatim}
{{Requirement}}
\end{verbatim}

Summary 1:
\begin{verbatim}
{{Summary 1}}
\end{verbatim}

Summary 2:
\begin{verbatim}
{{Summary 2}}
\end{verbatim}

Please provide your response.
\end{quote}

\noindent (3) \textbf{Prompt template for LLMEval}.

\begin{quote}
    In this task, you will be provided with a news article, a specific summary requirement, and a summary.

Your task is to rate the overall quality of the summary with a score from 1 to 5 concerning the summary requirement, where 1 is the lowest and 5 is the highest.

Please make sure you read and understand these instructions carefully. Please keep this document open while reviewing, and refer to it as needed.

Example Response:

Evaluation Form (scores ONLY):

- Overall Quality (1-5): 3

Here are the actual article, the summary requirement, and the summary:

Article:
\begin{verbatim}
{{Article}}
\end{verbatim}

Summary Requirement:
\begin{verbatim}
{{Requirement}}
\end{verbatim}

Summary:
\begin{verbatim}
{{SUMMARY}}
\end{verbatim}

Evaluation Form (scores ONLY):

- Overall Quality (1-5):
\end{quote}

\noindent (4) \textbf{Prompt template for LLMScore}.
\begin{quote}
    Answer the question based on the following article, a specific summary requirement, and a summary.
    
Question: Is the summary of good overall quality in relation to both the article and the summary requirement? 
(a). Yes. (b). No.

Article:
\begin{verbatim}
{{Article}}
\end{verbatim}

Summary Requirement:
\begin{verbatim}
{{Requirement}}
\end{verbatim}

Summary:
\begin{verbatim}
{{SUMMARY}}
\end{verbatim}

Answer: Yes
\end{quote}

\section{Crowd-Annotation Details}
\label{sec:appendix-interface}
\begin{figure*}[ht]
    \centering
    \includegraphics[width=0.9\linewidth]{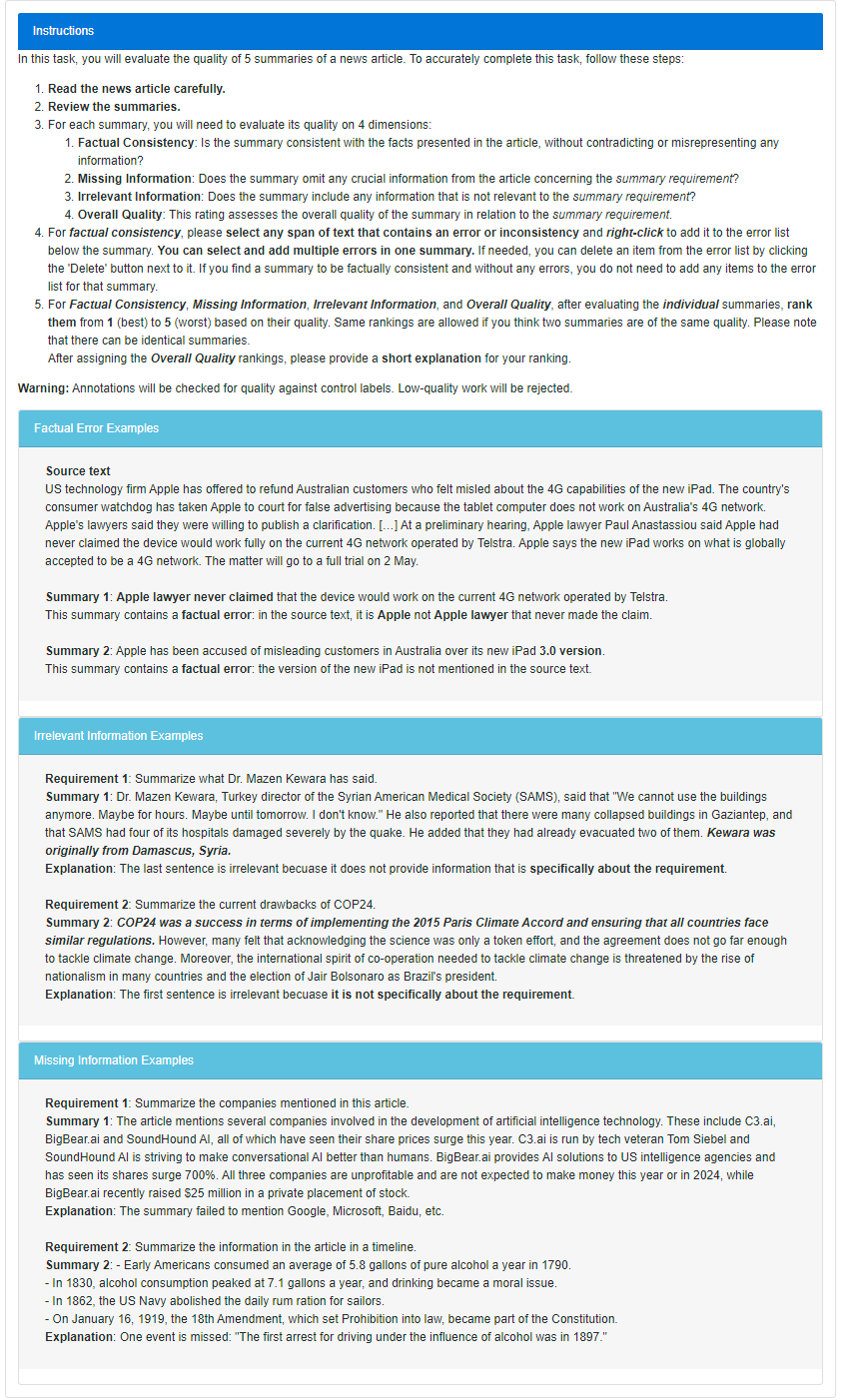}
 \caption{\label{fig:interface1}Annotation Interface Part 1: Instructions.
    }
\end{figure*}

\begin{figure*}[ht]
    \centering
    \includegraphics[width=1.0\linewidth]{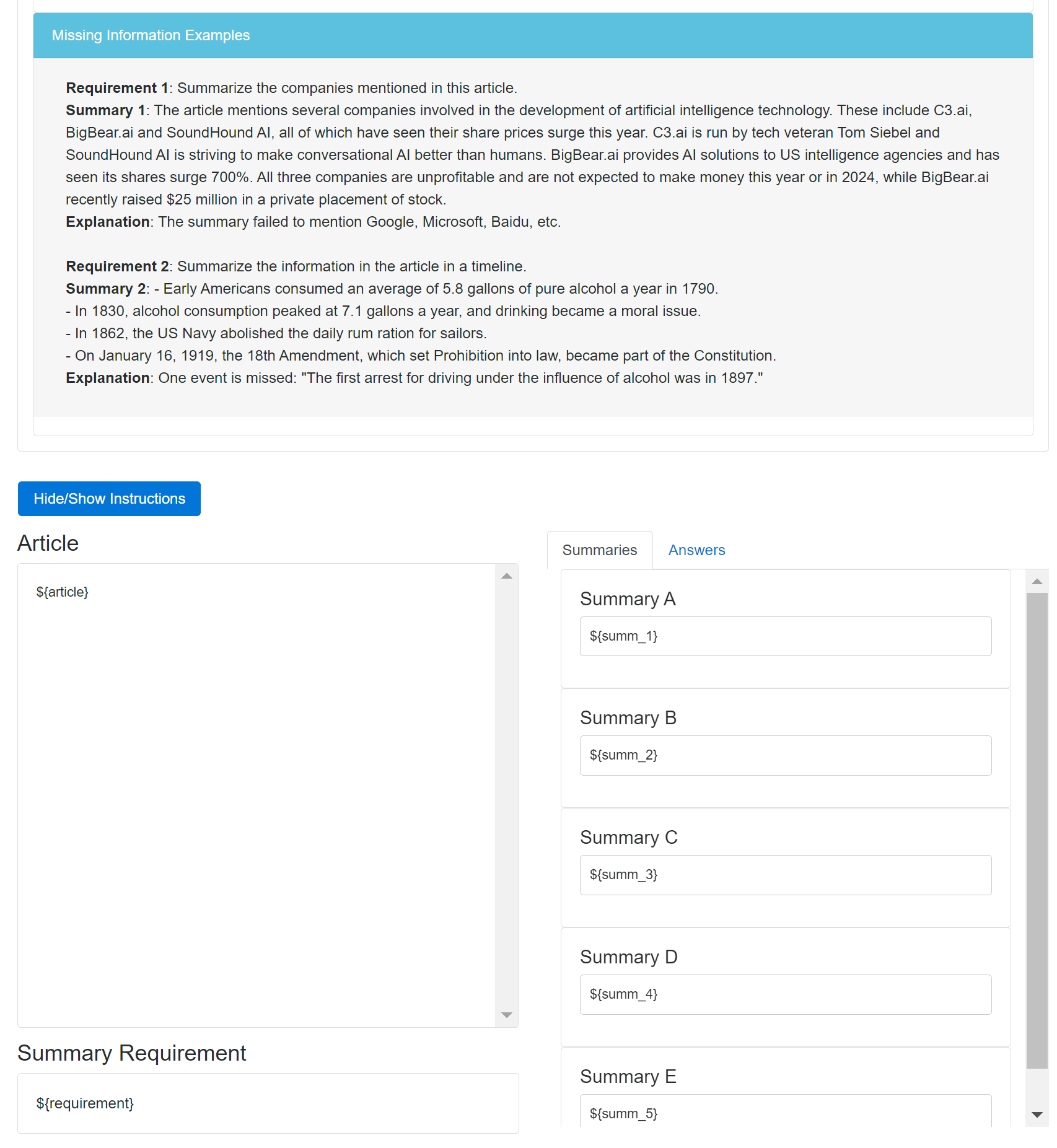}
 \caption{\label{fig:interface2}Annotation Interface Part 2: Data Input.
    }
\end{figure*}

\begin{figure*}[ht]
    \centering
    \includegraphics[width=1.0\linewidth]{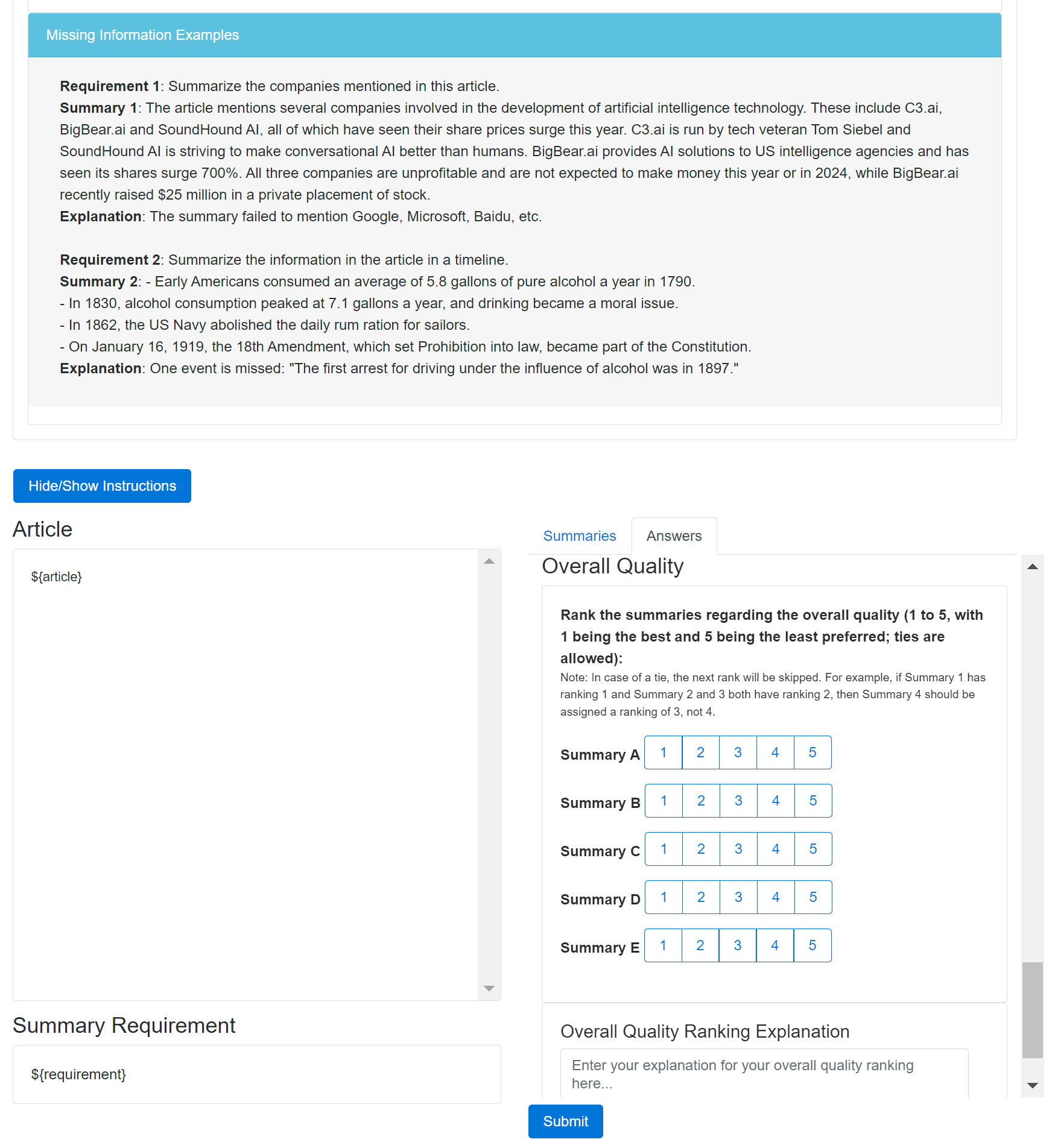}
 \caption{\label{fig:interface3}Annotation Interface Part 3: Result Collection.
    }
\end{figure*}

We provide screenshots of the human annotation interface we used for crowd-sourced summary evaluation (\S\ref{subsec:sys-human-eval}) in Figure \ref{fig:interface1}, \ref{fig:interface2}, and \ref{fig:interface3}.
We recruit MTurk annotators who are located in the US or the UK.
We set a competitive payment rate for better annotator retention, and the average hourly salary is around 20 US dollars.

\clearpage

\clearpage

\section{OpenAI's Model Index}
\label{sec:appendix-model-index}

Here we describe the training methods of OpenAI models we benchmarked for ins-controllable summarization using human evaluation (\S\ref{subsec:sys-human-eval}).
The following information was obtained from a blog post on the OpenAI's website, ``Model index for researchers.''\footnote{The original page is no longer accessible as of April 1st, 2024. An old snapshot of the page is available at \url{https://archive.ph/n5xMq}.}

\texttt{text-davinci-002}: Supervised fine-tuning (FeedME) on human-written demonstrations and on model samples rated 7/7 by human labelers on an overall quality score.

\texttt{text-davinci-003}: Reinforcement learning (PPO) with reward models trained from comparisons by humans.

The information about newer models can be found in \url{https://platform.openai.com/docs/models/}.

\section{Fine-grained Analysis of Hybrid LLM-Human Summaries}
\label{sec:appendix-human-eval-results}

\begin{table}
    \centering
    \small
    \begin{tabular}{lrrrr}
\toprule
  &   \textbf{Overall} &   \textbf{Missing} &   \textbf{Irrelevant} &   \textbf{Factual} \\
\midrule
 Tie      &      0.45 &      0.47 &         0.50 &      0.90 \\
 \midrule
 Initial     &      \textbf{0.30} &      \textbf{0.29} &         0.20 &      0.03 \\
 Hybrid   &      0.25 &      0.24 &         \textbf{0.30} &      \textbf{0.07} \\
\bottomrule
\end{tabular}
    \caption{Pairwise comparison between the initial LLM summary and the hybrid summary. Winning rates of both summaries are reported. 37\% of summaries are identical because no edits are made.}
    \label{tab:hybrid-eval}
\end{table}

In \S\ref{subsec:human-eval-results}, we found that the hybrid LLM-human summaries can not outperform the GPT-4 summaries on the \textit{overall quality} and \textit{missing information} dimensions.
To better understand the performance of the hybrid LLM-human summary, we use the obtained human annotations to perform a pairwise comparison between the initial LLM summary and the hybrid summary.
Results in Table \ref{tab:hybrid-eval} show that the hybrid summaries are better at the irrelevant information and factual consistency dimensions while worse at the overall quality and missing information dimensions.
We believe this is mainly because there are more ``delete'' than ``add'' editing operations in the hybrid summaries since we found that the initial LLM summaries are more likely to include irrelevant information than missing relevant information.
As a result, the reduced length of hybrid summaries may make them less favorable than the original summaries on the overall quality and missing information dimensions.

\section{Factual Error Examples}
\label{sec:appendix-fact-error}

We found that a considerable portion of the factual errors flagged by the crowd annotators is quite nuanced (\S\ref{subsec:sys-human-eval}).
Below we present a few examples.

\noindent \textbf{Example 1}

\begin{itemize}
    \item \textbf{Article (part)}: ... However, in 2007, Australian-listed firm Scotgold Resources entered the scene and revived the mine. It has been a rollercoaster ride ever since. \textbf{By 2013}, Scotgold had obtained planning permission and put a funding plan in place, only for the gold price to collapse, making the project less palatable for potential investors ...
    \item \textbf{Summary Error Span (with context)}: Planning permission was obtained \textbf{in 2013}.
\end{itemize}

\noindent \textbf{Example 2}

\begin{itemize}
    \item \textbf{Article (part)}: ...  Their efforts to attract backers were also undermined by the volatility of the precious metals market, which often saw gold prices slump. \textbf{By 2006}, the mine had changed hands several times and was up for sale once more. ...
    \item \textbf{Summary Error Span (with context)}: \textbf{2006}: Mine changed hands several times and was up for sale again.
\end{itemize}

\noindent \textbf{Example 3}

\begin{itemize}
    \item \textbf{Article (part)}: ...  There was also a divide when it came to trusting information on the virus coming from the WHO. About \textbf{one-third} of Republicans said they trusted WHO information versus 80\% of Democrats.  ...
    \item \textbf{Summary Error Span (with context)}: 80\% of Democrats trust information from the WHO, while only \textbf{33\%} of Republicans do.
\end{itemize}

\noindent \textbf{Example 4}

\begin{itemize}
    \item \textbf{Article (part)}: ...  Prof Suzuki says his team will need to carry on their work for many more years to be sure that the children of Fukushima are in the clear. But he and other experts now say they think there will be \textbf{very few, or even zero, extra childhood cancers} because of Fukushima. ...
    \item \textbf{Summary Error Span (with context)}: Professor Suzuki believes that the cases of thyroid cancer in Fukushima are not related to the nuclear disaster, and that the children of Fukushima \textbf{are not at risk} of developing cancer from the exposure to radiation. 
\end{itemize}

\noindent \textbf{Example 5}

\begin{itemize}
    \item \textbf{Article (part)}: ... Chum Mey says he was tortured, as his interrogators \textbf{tried to make him confess to spying for the US and Russia}. ... Eventually he said he \textbf{confessed to anything} so that the torture would be over. In his confession Chum Mey wrote that \textbf{he was working for the CIA} and had recruited dozens of agents in Cambodia. ... 
    \item \textbf{Summary Error Span (with context)}: Chum Mey was tortured until he eventually confessed to spying for the US and \textbf{Russia}. 
\end{itemize}

\section{Detailed Evaluation Settings of LLM-based Evaluation Methods}
\label{sec:appendix-meta-eval}
In \S\ref{sec:meta-eval}, we benchmark different LLM-based evaluation methods.
To this end, we use both system-level and summary-level correlations to evaluate the alignment between human evaluation and LLM-based evaluation. 
Specifically, given a correlation measurement function $\mathcal{M}$, e.g., the Kendall rank correlation coefficient~\citep{kendall1938new}, and two lists of average system scores $\bar{S}^{(f)}$ and $\bar{S}^{(h)}$ assigned by two evaluation methods $f$ and $h$, e.g., human evaluation and LLM evaluation, the system level correlation $\mathcal{C}_{\mathrm{sys}}$ between $f$ and $h$ is
\begin{equation}
\label{eq:sys-corr}
\mathcal{C}_{\mathrm{sys}} = \mathcal{M}(\bar{S}^{(f)}, \bar{S}^{(h)}).
\end{equation}
Similarly, the summary-level correlation $\mathcal{C}_{\mathrm{summ}}$ is an average of the correlation between two lists of scores, $S^{(f)}_i$ and $S^{(h)}_i$, assigned by the evaluation methods $f$ and $h$ for the summaries generated by different systems on each data example:
\begin{equation}
\label{eq:summ-corr}
\mathcal{C}_{\mathrm{summ}} = \sum_{i=0}^{N-1} \frac{\mathcal{M}(S^{(f)}_i, S^{(h)}_i)}{N},
\end{equation}
where $N$ is the size of the evaluation dataset.

Since we adopted a ranking-based evaluation protocol for our human evaluation collection (\S\ref{subsec:sys-human-eval}), we use the Kendall rank correlation coefficient as the correlation measurement.
Furthermore, apart from LLMRank, which directly generates a similar ranking, we convert the evaluation results of the other protocols to a ranking of different systems.
For LLMScore and LLMEval that perform direct scoring of summaries, we simply convert the scores into a ranking (ties are allowed).
For LLMCompare, we use the following scoring mechanism: (1) the winner system in a pairwise comparison receives 2 points, while the lost system receives 0 points;
(2) if there is a tie between two systems, each of them receives 1 point;
(3) the points are aggregated into a system ranking.

To remove the potential positional biases of the LLM-based evaluation methods~\citep{wang2023large, koo2023benchmarking}, we randomly shuffled the summary order when LLMRank or LLMCompare is used as the evaluation protocol.

\section{Length Bias in LLM-based Evaluation}
\label{sec:appendix-bias}

\begin{table*}[tbh!]
    \centering
\small
    \begin{tabular}{@{\extracolsep{1pt}}lcccccccccrr@{}}
\toprule
& \multicolumn{3}{c}{\textbf{Overall}} & \multicolumn{3}{c}{\textbf{Missing}} & \multicolumn{3}{c}{\textbf{Irrelevant}} &  \multicolumn{2}{c}{\textbf{Length}} \\
&\textbf{Win}&\textbf{Tie}&\textbf{Loss}&\textbf{Win}&\textbf{Tie}&\textbf{Loss}& \textbf{Win}&\textbf{Tie}&\textbf{Loss}&\textbf{System}&\textbf{GPT4}\\
 \cmidrule{2-4} \cmidrule{5-7} \cmidrule{8-10} \cmidrule{11-12} 
 gpt-3.5-turbo-0301     &  0.23 &  0.18 &   0.59 &  0.18 &  0.27 &   0.55 &  0.09 &  0.32 &   0.59 &     118.0 &     117.5 \\
 gpt-3.5-turbo-0613     &  0.15 &  0.10 &   0.75 &  0.00 &  0.45 &   0.55 &  0.00 &  0.60 &   0.40 &     110.7 &     111.8 \\
 gpt-3.5-turbo-instruct &  0.10 &  0.05 &   0.85 &  0.05 &  0.40 &   0.55 &  0.10 &  0.55 &   0.35 &     109.4 &     110.2 \\
 text-davinci-002       &  0.14 &  0.14 &   0.71 &  0.14 &  0.29 &   0.57 &  0.05 &  0.29 &   0.67 &      87.8 &      92.1 \\
 text-davinci-003       &  0.29 &  0.29 &   0.43 &  0.14 &  0.76 &   0.10 &  0.00 &  0.76 &   0.24 &      99.4 &      98.2 \\
 llama-2-7b-chat             &  0.38 &  0.10 &   0.52 &  0.24 &  0.29 &   0.48 &  0.24 &  0.33 &   0.43 &     132.1 &     115.1 \\
 llama-2-13b-chat            &  0.29 &  0.19 &   0.52 &  0.14 &  0.52 &   0.33 &  0.10 &  0.48 &   0.43 &     113.6 &     110.2 \\
 llama-2-70b-chat            &  0.30 &  0.20 &   0.50 &  0.25 &  0.55 &   0.20 &  0.25 &  0.50 &   0.25 &     110.8 &     102.5 \\
 mistral-instruct            &  0.20 &  0.15 &   0.65 &  0.20 &  0.20 &   0.60 &  0.10 &  0.25 &   0.65 &      91.3 &      90.0 \\
 gpt-4-1106-preview     &  0.50 &  0.20 &   0.30 &  0.45 &  0.40 &   0.15 &  0.30 &  0.60 &   0.10 &      98.75 &      91.4 \\
\bottomrule
\end{tabular}
    \caption{Automatic benchmarking results on summary pairs with similar lengths. The summaries of different LLMs are compared against summaries generated by GPT-4 (\texttt{gpt-4-0314}) using the LLMCompare protocol powered by \texttt{gpt-4-0314}. The number of wins, ties, and losses is reported as well as the average summary length.
    }
    \label{tab:auto-eval-length}
\end{table*}

As a further investigation of \S\ref{subsec:meta-eval-result}, we conducted a case study by only comparing summary pairs with similar lengths.
Specifically, for each pair of systems, we keep only those pairs of summaries where the difference in lengths falls within the 20th percentile.
The results in Table~\ref{tab:auto-eval-length} indicate that when the length difference is controlled, none of the LLMs we compared can outperform GPT-4 on the overall quality dimension, and Llama-2 models no longer have a clear advantage over GPT-4. 
We note that since different examples are used for the comparison of different system pairs, the results in Table~\ref{tab:auto-eval-length} are no longer directly comparable.

\section{Comparing Generic and Requirement-Specific Summaries}

\begin{table}
    \centering
    \small
\addtolength{\tabcolsep}{-3pt} 
    \begin{tabular}{lrrrr}
\toprule
  \textbf{Model} &   \textbf{R1} &   \textbf{R2} &   \textbf{Specific} & \textbf{Generic} \\
\midrule
 \texttt{gpt-3.5-turbo-0301} & 47.21 & 26.30 & 127.3 & 144.0 \\
 \texttt{gpt-4-0314} & 43.00 & 19.37 & 117.1 & 123.7 \\
\bottomrule
\end{tabular}
\addtolength{\tabcolsep}{3pt} 
    \caption{The similarities between the generic and requirement-specific summaries as measured in ROUGE-1/2 (R1/R2). The average summary length is also reported, denoted by \texttt{Specific} and \texttt{Generic} respectively.}
    \label{tab:generic}
\end{table}

\begin{table}
    \centering
    \small
\addtolength{\tabcolsep}{-3pt} 
    \begin{tabular}{lrrrr}
\toprule
  \textbf{Model} &   \textbf{R1} &   \textbf{R2} &   \textbf{Greedy} & \textbf{Sampled} \\
\midrule
 \texttt{gpt-3.5-turbo-0301} & 61.63 & 38.11 & 127.3 & 125.48 \\
 \texttt{gpt-4-0314} & 66.16 & 43.20 & 117.1 & 117.7 \\
\bottomrule
\end{tabular}
\addtolength{\tabcolsep}{3pt} 
    \caption{The similarities between the greedy-decoded and sampled requirement-specific summaries as measured in ROUGE-1/2 (R1/R2). The average summary length is denoted by \texttt{Greedy} and \texttt{Sampled} respectively.}
    \label{tab:sampled}
\end{table}

As a case study, we use \texttt{gpt-3.5-turbo-0301} and \texttt{gpt-4-0314} to generate \textit{generic} summaries without specific requirements, and compare them with the requirement-specific summaries using \texttt{gpt-4-0314} as the evaluator with the LLMCompare protocol.
The evaluation results show that the requirement-specific summaries generated by \texttt{gpt-4-0314} have a winning rate of 97\% over the generic summaries on the \textit{overall quality} dimension, while those with \texttt{gpt-3.5-turbo-0301} have a winning rate of 96\%.
We also evaluate the similarity between the generic and requirement-specific summaries in Table~\ref{tab:generic}, as well as the summary length.
In addition, in Table~\ref{tab:sampled}, we report the similarity between greedy-decoded and sampled (with a temperate of 1.0) requirement-specific summaries.
Results in Table~\ref{tab:generic} and Table~\ref{tab:sampled} suggest that the similarity between the generic and requirement-specific summaries is relatively low, and the generic summaries are not preferred by \texttt{gpt-4-0314} despite its longer average length.

\section{\textsc{InstruSumEval} Leaderboard}
\label{appendix:instrusumeval}

To facilitate future studies in developing reliable automatic evaluation methods for ins-controllable summarization, we expand the experiments in \S\ref{sec:meta-eval} and \S\ref{sec:benchmark} to collect a large-scale leaderboard of automatic evaluators using the human annotations we collected in this work.

\subsection{Evaluation Setting}
We note that our meta-evaluation setting on ins-controllable summarization can be viewed as a special case of meta-evaluation of instruction-following evaluation~\cite{zheng2023judging, dubois2023alpacafarm, zeng2024evaluating}.
Therefore, to align with the evaluation setting in these works, we focus on evaluating various LLMs at performing \textit{pairwise} comparisons.
This setting is similar to the LLMCompare protocol we investigated in \S\ref{sec:meta-eval}.
However, we perform a further data transformation step, combining the source article and the summary requirement as a complex \textit{instruction}, to match the task setting of instruction-following.
We note the difference between \textsc{InstruSumEval} and the existing meta-evaluation benchmarks for instruction-following evaluation such as LLMBar~\citep{zeng2024evaluating} -- the instructions in \textsc{InstruSumEval} contain more than 1000 words on average, while LLMBar's instructions contain around 50 words on average.
Therefore, \textsc{InstruSumEval} can be viewed as a long-context meta-evaluation benchmark for instruction-following evaluation.

We adopt the prompt template used in \citet{zeng2024evaluating} for the pairwise comparison, which incorporates several evaluation rules in the prompt and requires the LLMs to predict an answer directly:
 \begin{quote}
You are a helpful assistant in evaluating the quality of the outputs for a given instruction. Your goal is to select the best output for the given instruction.
\newline
\newline
\textbf{[User Message]}\\
Select the Output (a) or Output (b) that is better for the given instruction. The two outputs are generated by two different AI chatbots respectively. \\
\newline
Here are some rules of the evaluation: \\
(1) You should prioritize evaluating whether the output honestly/precisely/closely executes the instruction, then consider its helpfulness, accuracy, level of detail, harmlessness, etc. \\
(2) Outputs should NOT contain more/less than what the instruction asks for, as such outputs do NOT precisely execute the instruction.\\
(3) You should avoid any potential bias and your judgment should be as objective as possible. For example, the order in which the outputs were presented should NOT affect your judgment, as Output (a) and Output (b) are equally likely to be the better.
\newline
\newline
Do NOT provide any explanation for your choice.\\
Do NOT say both / neither are good.\\
You should answer using ONLY "Output (a)" or "Output (b)". Do NOT output any other words.\\
\newline

\# Instruction: \\
\{INSTRUCTION\}
\newline
\newline
\# Output (a): \\
\{OUTPUT\_1\}
\newline
\newline
\# Output (b): \\
\{OUTPUT\_2\}
\newline
\newline
\# Which is better, Output (a) or Output (b)? Your response should be either "Output (a)" or "Output (b)":
\end{quote}

To account for the potential inconsistency in the LLM-evaluators' decisions, we perform the evaluation in two directions by swapping the output (summary) pairs.
We then focus on evaluating two properties of the LLM-evaluators -- their \textit{evaluation accuracy} using human annotations and their \textit{self-consistency} in their bidirectional evaluation results. 
For both properties, we calculate the accuracy and the agreement rate in Krippendorff's alpha.

\subsection{Data Filtering}
To ensure the quality of human annotation, we perform a data filtering process.
Specifically, we first convert the ranking-based individual annotations to pairwise comparisons. 
With 100 ranking-based annotations and 5 systems, we obtain 1000 pairwise annotations. 
We then remove pairwise annotations without perfect human agreement and any ties, resulting in 411 data instances.

\subsection{Results}
Table~\ref{tab:instrusumeval} displays the performance of 41 LLMs we evaluated.
We find that there is a large performance gap between LLMs with different capacities.

\begin{table*}
    \centering
    \small
    \begin{tabular}{llrrrr}
\toprule
 \textbf{Index}   & \textbf{Model  }                & \textbf{Accuracy}       & \textbf{Agreement}      & \textbf{Self-Accuracy }  & \textbf{Self-Agreement}   \\
\midrule
 1       & \href{https://platform.openai.com/docs/models/gpt-4-turbo-and-gpt-4}{gpt-4-0613}             & \textbf{0.804} & \textbf{0.599} & 0.861           & 0.710            \\
 2       & \href{https://platform.openai.com/docs/models/gpt-4o}{gpt-4o}                                & 0.803          & 0.595          & 0.917           & \textbf{0.823}   \\
 3       & \href{https://platform.openai.com/docs/models/gpt-4-turbo-and-gpt-4}{gpt-4-0314}             & 0.790          & 0.570          & 0.788           & 0.571            \\
 4       & \href{https://huggingface.co/meta-llama/Meta-Llama-3-70B-Instruct}{llama-3-70b}              & 0.786          & 0.565          & 0.825           & 0.642            \\
 5       & \href{https://mistral.ai/news/mistral-large/}{mistral-large}                                 & 0.785          & 0.558          & 0.832           & 0.645            \\
 6       & \href{https://platform.openai.com/docs/models/gpt-4-turbo-and-gpt-4}{gpt-4-0125-preview}     & 0.782          & 0.554          & 0.837           & 0.659            \\
 7       & \href{https://platform.openai.com/docs/models/gpt-4-turbo-and-gpt-4}{gpt-4-turbo-2024-04-09} & 0.783          & 0.551          & 0.878           & 0.725            \\
 8       & \href{https://platform.openai.com/docs/models/gpt-4-turbo-and-gpt-4}{gpt-4-1106-preview}     & 0.779          & 0.548          & 0.796           & 0.579            \\
 9       & \href{https://docs.anthropic.com/en/docs/about-claude/models}{claude-3.5-sonnet}             & 0.775          & 0.540          & 0.803           & 0.599            \\
 10      & \href{https://deepmind.google/technologies/gemini/flash/}{gemini-1.5-flash}                  & 0.775          & 0.535          & 0.793           & 0.562            \\
 11      & \href{https://deepmind.google/technologies/gemini/pro/}{gemini-1.5-pro}                      & 0.757          & 0.503          & 0.757           & 0.519            \\
 12      & \href{https://docs.anthropic.com/en/docs/about-claude/models}{claude-3-opus}                 & 0.741          & 0.473          & 0.730           & 0.470            \\
 13      & \href{https://huggingface.co/Qwen/Qwen2-72B-Instruct}{qwen-2-72b}                            & 0.731          & 0.443          & 0.818           & 0.603            \\
 14      & \href{https://huggingface.co/THUDM/glm-4-9b-chat}{glm-4-9b}                                  & 0.734          & 0.440          & 0.808           & 0.518            \\
 15      & \href{https://huggingface.co/CohereForAI/c4ai-command-r-plus}{command-r-plus}                & 0.709          & 0.395          & 0.779           & 0.494            \\
 16      & \href{https://huggingface.co/Qwen/Qwen1.5-72B-Chat}{qwen-1.5-72b}                            & 0.692          & 0.387          & 0.618           & 0.235            \\
 17      & \href{https://deepmind.google/technologies/gemini/pro/}{gemini-1.0-pro}                      & 0.687          & 0.385          & 0.686           & 0.399            \\
 18      & \href{https://docs.anthropic.com/en/docs/about-claude/models}{claude-2.1}                    & 0.661          & 0.347          & 0.603           & 0.252            \\
 19      & \href{https://huggingface.co/mistralai/Mixtral-8x7B-Instruct-v0.1}{mixtral-8x7b}             & 0.687          & 0.344          & 0.774           & 0.439            \\
 20      & \href{https://huggingface.co/allenai/tulu-2-dpo-70b}{tulu-2-dpo-70b}                         & 0.661          & 0.337          & 0.603           & 0.273            \\
 21      & \href{https://huggingface.co/01-ai/Yi-1.5-34B-Chat}{yi-1.5-34b}                              & 0.669          & 0.330          & 0.474           & 0.173            \\
 22      & \href{https://huggingface.co/meta-llama/Llama-2-70b-chat-hf}{llama-2-70b}                    & 0.669          & 0.324          & 0.771           & 0.537            \\
 23      & \href{https://huggingface.co/allenai/tulu-2-70b}{tulu-2-70b}                                 & 0.647          & 0.300          & 0.582           & 0.262            \\
 24      & \href{https://huggingface.co/Qwen/Qwen1.5-32B-Chat}{qwen-1.5-32b}                            & 0.662          & 0.296          & 0.708           & 0.331            \\
 25      & \href{https://docs.anthropic.com/en/docs/about-claude/models}{claude-3-sonnet}               & 0.663          & 0.292          & 0.730           & 0.401            \\
 26      & \href{https://platform.openai.com/docs/models/gpt-3-5-turbo}{gpt-3.5-turbo-0125}             & 0.635          & 0.254          & 0.562           & 0.225            \\
 27      & \href{https://huggingface.co/01-ai/Yi-1.5-9B-Chat}{yi-1.5-9b}                                & 0.631          & 0.222          & 0.652           & 0.188            \\
 28      & \href{https://huggingface.co/allenai/tulu-2-13b}{tulu-2-13b}                                 & 0.618          & 0.219          & 0.669           & 0.319            \\
 29      & \href{https://huggingface.co/allenai/tulu-2-dpo-13b}{tulu-2-dpo-13b}                         & 0.612          & 0.210          & 0.642           & 0.270            \\
 30      & \href{https://huggingface.co/meta-llama/Llama-2-13b-chat-hf}{llama-2-13b}                    & 0.608          & 0.201          & 0.679           & 0.375            \\
 31      & \href{https://docs.anthropic.com/en/docs/about-claude/models}{claude-3-haiku}                & 0.628          & 0.200          & 0.727           & 0.188            \\
 32      & \href{https://docs.anthropic.com/en/docs/about-claude/models}{claude-instant-1.2}            & 0.605          & 0.187          & 0.569           & 0.222            \\
 33      & \href{https://huggingface.co/meta-llama/Meta-Llama-3-8B-Instruct}{llama-3-8b}                & 0.617          & 0.166          & 0.827           & 0.230            \\
 34      & \href{https://huggingface.co/mistralai/Mistral-7B-Instruct-v0.3}{mistral-7b-v0.3}            & 0.606          & 0.144          & 0.883           & 0.463            \\
 35      & \href{https://huggingface.co/mistralai/Mistral-7B-Instruct-v0.1}{mistral-7b-v0.1}            & 0.585          & 0.107          & 0.871           & 0.531            \\
 36      & \href{https://huggingface.co/allenai/tulu-2-dpo-7b}{tulu-2-dpo-7b}                           & 0.589          & 0.101          & 0.835           & 0.126            \\
 37      & \href{https://huggingface.co/mistralai/Mistral-7B-Instruct-v0.2}{mistral-7b-v0.2}            & 0.582          & 0.089          & 0.854           & 0.293            \\
 38      & \href{https://huggingface.co/google/gemma-7b-it}{gemma-7b}                                   & 0.574          & 0.069          & 0.859           & 0.207            \\
 39      & \href{https://huggingface.co/allenai/tulu-2-7b}{tulu-2-7b}                                   & 0.578          & 0.067          & \textbf{0.929}  & 0.202            \\
 40      & \href{https://huggingface.co/google/gemma-2b-it}{gemma-2b}                                   & 0.566          & 0.063          & 0.788           & 0.168            \\
 41      & \href{https://huggingface.co/meta-llama/Llama-2-7b-chat-hf}{llama-2-7b}                      & 0.564          & 0.048          & 0.873           & 0.264            \\
\bottomrule
\end{tabular}
    \caption{\textsc{InstruSumEval} leaderboard for LLMs' performance at ins-controllable summarization evaluation. \textbf{Accuracy} and \textbf{Agreement} are calculated against human annotations, \textbf{Self-Accuracy} and \textbf{Self-Agreement} are calculated between the LLMs' bidirectional predictions.}
    \label{tab:instrusumeval}
\end{table*}

\end{document}